\documentclass{bmvc2k}

\title{Recovering Global Data Distribution Locally in Federated Learning}

\addauthor{Ziyu Yao}{yaozy@stu.pku.edu.cn}{1}

\addinstitution{
 Peking University\\
 Beijing, China
}

\runninghead{Yao}{Recovering Global Data Distribution Locally in FL}

\def\eg{\emph{e.g}\bmvaOneDot}

\def\ie{\emph{i.e}\bmvaOneDot}

\usepackage[capitalize]{cleveref}
\crefname{section}{Sec.}{Secs.}
\Crefname{section}{Section}{Sections}
\Crefname{table}{Table}{Tables}
\crefname{table}{Tab.}{Tabs.}

\usepackage{graphicx}
\usepackage{amsmath}
\usepackage{amssymb}
\usepackage{booktabs}
\usepackage{color}
\usepackage{nicefrac}
\usepackage[ruled,vlined,linesnumbered]{algorithm2e}
\usepackage{algorithmicx}
\usepackage{listings}
\usepackage{etoolbox}
\usepackage{xpatch}
\usepackage{array}
\usepackage{multirow}
\usepackage{graphicx}
\usepackage{colortbl}
\usepackage{pifont}
\usepackage{enumitem}
\usepackage{xcolor}
\usepackage{bm}
\usepackage{wrapfig,lipsum,booktabs}

\begin{document}

\maketitle

\begin{abstract}

Federated Learning (FL) is a distributed machine learning paradigm that enables collaboration among multiple clients to train a shared model without sharing raw data. However, a major challenge in FL is the label imbalance, where clients may exclusively possess certain classes while having numerous minority and missing classes. Previous works focus on optimizing local updates or global aggregation but ignore the underlying imbalanced label distribution across clients. In this paper, we propose a novel approach \textbf{ReGL} to address this challenge, whose key idea is to \textbf{Re}cover the \textbf{G}lobal data distribution \textbf{L}ocally. Specifically, each client uses generative models to synthesize images that complement the minority and missing classes, thereby alleviating label imbalance. Moreover, we adaptively fine-tune the image generation process using local real data, which makes the synthetic images align more closely with the global distribution. Importantly, both the generation and fine-tuning processes are conducted at the client-side without leaking data privacy. Through comprehensive experiments on various image classification datasets, we demonstrate the remarkable superiority of our approach over existing state-of-the-art works in fundamentally tackling label imbalance in FL.

\end{abstract}
\section{Introduction}

Federated Learning (FL) is a distributed machine learning paradigm that allows multiple devices or clients to collaboratively train a shared model without directly exchanging raw data. In the context of FL, each device only needs to send its model parameter updates (\eg, gradients) to a central server for integration, without sharing the local data. Consequently, the raw data remain at their original location, reducing the risk of data leakage or misuse. Recently, the widespread use of digital devices and mobile computing, along with the rapid development of deep learning~\cite{he2016deep, yao2024building, yao2024multi}, has led to a surge in research and applications of FL. Particularly, in healthcare~\cite{feki2021federated, li2019privacy, oldenhof2023industry}, finance~\cite{feng2022data, li2020blockchain, schreyer2022federated}, and the Internet of Things~\cite{chiu2020semisupervised, zhao2021federated, lu2019blockchain}, FL has shown remarkable potential.

As for traditional centralized training, one can easily balance the dataset. However, in FL, the data exhibits non-IID (Independent and Identically Distributed) characteristics due to its decentralized nature. This can arise from the varied data collections at different nodes or the inherent differences in local data characteristics, thereby posing a challenge in FL. In many real-world scenarios, the \textbf{label distribution skew} is one of the most pronounced manifestations of this non-IID data distribution, resulting in certain labels being over-represented in some clients while being under-represented in others~\cite{zhao2018federated, li2022federated, zhang2022federated}. This skewness can severely degrade the performance, leading to a necessity for addressing label imbalance.

\begin{wrapfigure}{r}{0.55\textwidth}
\begin{center}
\vskip -0.1in
\centerline{\includegraphics[width=0.55\columnwidth]{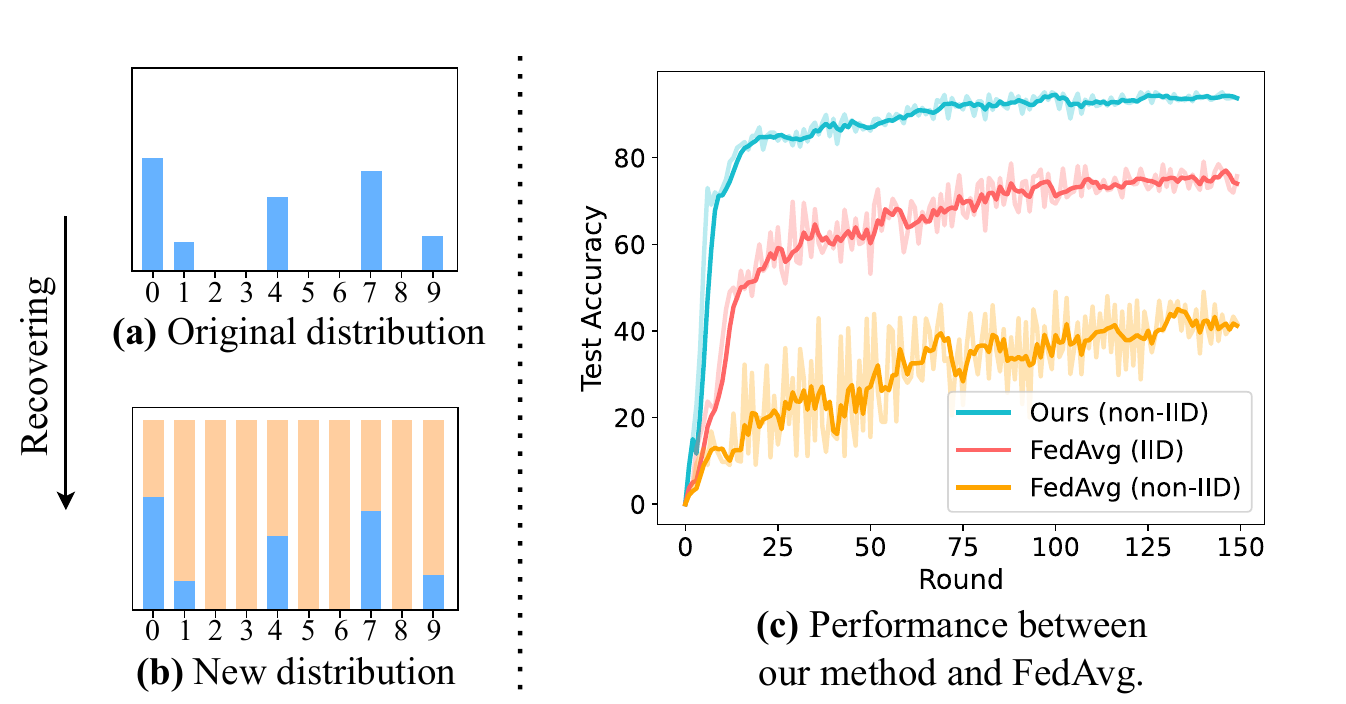}}
\caption{{\bf (a)} Imbalanced label distribution for one client. {\bf (b)} Recovered data distribution at client-side. The \textcolor[RGB]{51, 153, 255}{blue} and \textcolor[RGB]{255, 153, 51}{orange} histogram represent the number of real and synthetic images. {\bf (c)} The comparison between our method and FedAvg under both IID and non-IID settings.}
\label{fig:introduction}
\end{center}
\vskip -0.3in
\end{wrapfigure}

Currently, various techniques have been proposed to address label imbalance in centralized settings, such as oversampling~\cite{he2008adasyn, han2005borderline, chawla2002smote}, undersampling~\cite{triguero2016evolutionary, kim2016optimization, popel2018hybrid}, or generating synthetic data~\cite{wang2020deep, mullick2019generative, douzas2018effective}, however, their direct applicability in FL is trivial due to data privacy constraints and decentralized data ownership. Particularly under extreme distributions, clients may directly lack samples of a certain class, rendering traditional imbalanced learning methods unable to handle the scenario where the number of samples for a certain class is zero. Moreover, many approaches in FL have mainly focused on mitigating label imbalance by optimizing local updates~\cite{li2020federated} or global aggregation~\cite{wang2020tackling}. However, they have not addressed the underlying issue of label imbalance, \ie, \textbf{the significant misalignment between the local distribution of clients and the global distribution}, making these approaches always suboptimal when dealing with label distribution skew. Drawing from the dilemmas presented in previous works, we intuitively pose the following question:

\textbf{Question}: \textit{If the data distribution among clients in FL is balanced, can the performance degradation caused by label imbalance be fundamentally addressed?} 

Taking a 10-category subset~\cite{russakovsky2015imagenet} of ImageNet as an example, to answer this question, we first demonstrate a typical label distribution of a random client in~\cref{fig:introduction} (a). It is evident that the label distribution is severely biased, leading to overfitting on majority classes and difficulty in accurately identifying missing or minority classes.
To address this issue, we propose a novel method to recover the biased distribution, ensuring label balance for each client (details will be discussed in subsequent sections). 
In~\cref{fig:introduction} (b), we visualize the label distribution after applying our method. It can be observed that the distribution is now more balanced compared to the initial distribution shown in~\cref{fig:introduction} (a). To evaluate the performance of our method, we conduct experiments using the recovered balanced data in~\cref{fig:introduction} (c), and our method outperforms FedAvg in both non-IID and IID scenarios. Even in IID scenarios where FedAvg is designed to perform well, our method surpasses it by 15\% in terms of accuracy. These results validate that the balanced local distribution contributes to the global performance. So the goal of this paper is to \textit{recover each imbalanced local distribution back to the balanced global distribution without compromising the privacy of other clients}.

To achieve this goal, we propose a novel approach named \textbf{ReGL}, aimed at \textbf{Re}covering \textbf{G}lobal data distribution \textbf{L}ocally. We evaluate our paradigm on image classification task. In detail, each client initially possesses a specific quantity of real images that conform to the label distribution skew. Subsequently, we propose a training-free approach, where clients directly use foundation generative models~\cite{rombach2022high} to generate synthetic images for each class. Following this step, clients combine both real and synthetic data to train their local models. By incorporating synthetic data, the missing and minority classes can be supplemented, thus alleviating the label imbalance. However, different datasets exist with domain gaps, such as painting or realistic styles, which make synthetic data insufficiently aligned with the global distribution. Therefore, we propose that clients employ adaptive approaches~\cite{hu2021lora} to fine-tune their generative models locally based on their real data, thus privacy is preserved. This process enhances better adaptation to specific data distributions for each dataset, with the goal of effectively recovering the global data distribution at the client-side.

The contribution can be summarized as follows:
\begin{itemize}
    \item Our proposed ReGL aims to harness the power of foundation generative models to supplement the minority and missing classes for each client, thus alleviating the label imbalance problem without violating privacy constraints.
    \item By fine-tuning the foundation generative models using specific local data at the client-side, the synthetic data can be much more representative to the global data distribution. This fundamentally recovers the global data distribution locally to effectively addresses the label distribution skew challenge.
    \item We conduct comprehensive experiments to evaluate the effectiveness of our proposed ReGL across various datasets, which exhibits remarkable superiority over existing state-of-the-art FL algorithms on both global generalization and local personalization, with an average increase of 30\%.
\end{itemize}

\section{Preliminary and Related Work}

\paragraph{FL with Non-IID Data} While the meaning of IID is generally clear, data can exhibit non-IID characteristics in various ways, such as feature distribution skew, label distribution skew, and quantity skew, etc~\cite{yu2020federated, wang2020federated, kairouz2021advances}. Different non-IID regimes may require the development of distinct mitigation strategies. As observed in~\cite{li2022federated}, there is no single studied algorithm that consistently exhibits good performance across all non-IID settings. Therefore, our work aims to address the challenges specifically associated with label distribution skew. 

\paragraph{Label Distribution Skew} Let us consider the global distribution denoted as $P_{\text{global}}(x, y)$, where $x$ represents the input data and $y$ denotes the corresponding label. Additionally, we have the local distributions of client $i$, denoted as $P_i(x, y)$. Each client $i$ is capable of sampling a data point $(x, y)$ from its local distribution, \ie, $(x, y) \sim P_i(x, y)$. These distributions can be expressed as the product of conditional probabilities $P_i(x \mid y)$ and marginal probabilities $P_i(y)$, yielding $P_i(x\mid y)P_i(y)$. 
In label distribution skew, the marginal probability $P_i(y)$ may vary across different clients while $P_i(x \mid y)$ remains the same, resulting in local distributions that deviate significantly from the global distribution $P_{\text{global}}(x, y)$. We show the illustration of label distribution skew among clients in~\cref{fig:partition}.

\paragraph{FL with Label Distribution Skew} Recent works that address label distribution skew can be categorized into 1) incorporating momentum and adaptive methods~\cite{reddi2020adaptive}, 2) reducing bias in local model updates~\cite{karimireddy2020scaffold}, 3) regularizing local objective functions~\cite{li2020federated}, and 4) considering alternative aggregation methods~\cite{wang2020tackling}. We discuss them in detail in~\cref{sec:detail_related_work}.

However, they are still suboptimal as they cannot address the misalignment between local and global data distribution. In contrast, our work starts by recovering the global distribution at the client-side, which can fundamentally address the label distribution skew.

\paragraph{Foundation Generative Model} Foundation generative models, such as Stable Diffusion~\cite{rombach2022high}, DALL-E2~\cite{ramesh2022hierarchical}, Imagen~\cite{saharia2022photorealistic}, and GLIDE~\cite{nichol2021glide}, showcase impressive abilities in generating images based on text prompts. These models are based on the diffusion models, which are formulated as time-conditional denoising networks that learn the reverse process of a Markov Chain. Recently, Low-Rank Adaptation (LoRA)~\cite{hu2021lora} has been proposed to fine-tune foundation generative models, allowing for customization of the generated images.
\section{Problem Setup and Motivation}

\paragraph{Problem Setting} Suppose there are $M$ clients, each possessing their own private datasets $\mathcal{D}_m = \{(x_i, y_i)\}^{N_m}_{i=1}$, where $x_i$ represents the image, $y_i$ corresponds to its associated label, and $N_m$ is the number of samples on the $m$-th client. In label distribution skew, the label sets $\mathcal{Y}_m$ of different clients vary. 
We evaluate our approach on both \textbf{generalization} and \textbf{personalization} aspects. 
\textbf{1) Global generalization task:} Our goal is to learn a model parameterized with $\vartheta$ over the dataset $\mathcal{D} = \bigcup_{m=1}^{M} \mathcal{D}_m$ in the server without access to the original data:
\begin{equation}
\mathop{\min}\limits_{\vartheta}\sum_{m=1}^M \mathbb{E}_{(x, y)\thicksim \mathcal{D}_m}[l_m(\vartheta;(x, y))],
\end{equation}
where $l_m$ is the loss function for the $m$-th client.
\textbf{2) Local personalization task:} Our goal is to learn the personalized model $\vartheta_m$ for each client $m$ that performs optimally on their respective
local dataset $\mathcal{D}_m$. The objective is to minimize the loss:
\begin{equation}
\mathop{\min}\limits_{\vartheta_1, \vartheta_2, \dots, \vartheta_M}\sum_{m=1}^M \mathbb{E}_{(x, y)\thicksim \mathcal{D}_m}[l_m(\vartheta_m;(x, y))].
\end{equation}

\paragraph{Motivation} In label distribution skew, due to stringent privacy requirements, clients operate in isolation, preventing them from gaining insights into the distribution of other clients. As a result, the global distribution remains obfuscated, and label imbalance challenges persist. Consider the following assumption: \textit{If the local distribution $P_i(x, y)$ aligns with global distribution $P_{\text{global}}(x, y)$, then the influence of label distribution skew should be mitigated.}

\begin{wraptable}{R}{6.3cm}
\fontsize{8pt}{8pt}\selectfont
\begin{center}
\vskip -0.25in
\begin{tabular}{>{\centering\arraybackslash}m{2.1cm} | >{\centering\arraybackslash}m{0.36cm}>{\centering\arraybackslash}m{0.36cm}>{\centering\arraybackslash}m{0.36cm}>{\centering\arraybackslash}m{0.36cm}>{\centering\arraybackslash}m{0.36cm}}
    \toprule
    Global Proportion & 0\%$^{\text{\textdagger}}$ & 5\% & 10\% & 20\% & 30\% \\
    \midrule
    Generalization & 36.3 & 39.1 & 43.1 & 50.8 & {\bf 55.8} \\
    Personalization & 38.6 & 40.2 & 44.6 & 51.6 & {\bf 57.3} \\
    \bottomrule
\end{tabular}
\end{center}
\vskip -0.08in
\caption{Comparison of different proportions of global data allocated to each client. ${\text{\textdagger}}$: Completely non-IID setting.}
\label{tab:motivation}
\vskip -0.1in
\end{wraptable}

To validate this assumption, we collect entire data from all clients on the server side and sample different proportions from this balanced dataset to distribute to each client, thus alleviating local data imbalance. \textit{Note that this operation completely violates privacy and is solely used for analysis purposes.} 
To be specific, we use the Dirichlet distribution~\cite{li2021model} with $\beta=0.01$ to simulate label distribution skew on ImageFruit~\cite{zhang2023federated} dataset across clients (details can be found in~\cref{sec:setup}), and then we allocate an equal proportion of data from each label of the global dataset to each client, varying the percentage from 5\% to 30\%. 
This allocation, sourced from the global dataset, maintains uniformity within each label, thereby the local distribution of each client partially recovers the global distribution. 

As depicted in~\cref{tab:motivation}, when global data accessibility is absent, FedAvg is trained under a wholly non-IID scenario, resulting in poor accuracies. This starkly emphasizes the substantial repercussions of label distribution skew. Conversely, with an increasing proportion of global data allocation, the accuracy exhibits remarkable enhancement. Specifically, when 30\% of the global data is allocated to clients, the final accuracy reaches 55.8\% and 57.3\% for two tasks. This inspire us to consider: \textit{How can we recover the global data distribution using only local data information without compromising the privacy of other clients?}
\section{Recovering Global Distribution Locally}

\begin{figure*}[t]
  \centering
  \includegraphics[width=0.9\linewidth]{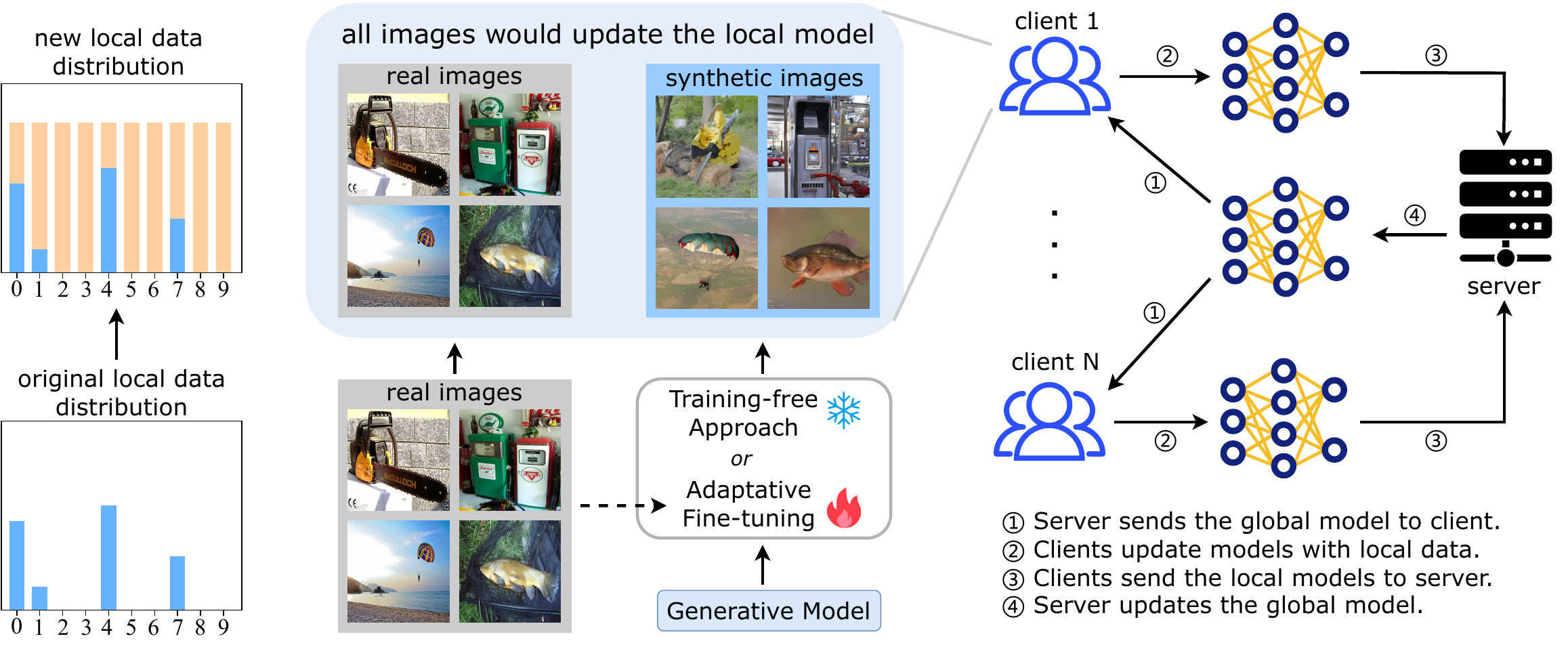}
  \caption{Our proposed ReGL framework. We use generative models to generate data at the client-side, thereby alleviating label imbalance. To better recover the global distribution, clients fine-tune their generative models using local data. Both real and synthetic data would be used to update the local models, resulting in a more balanced global aggregation model.}
  \label{fig:method}
\end{figure*}

\paragraph{Framework Overview} As shown in~\cref{fig:method}, our ReGL leverages generative models at the client-side, aiming to recover the global distribution locally. Considering the computational limitations, we propose two approaches: training-free and fine-tuning. In training-free approach, we use Stable Diffusion (SD)~\cite{rombach2022high} to generate data for each client, thereby complementing the minority and missing classes. However, different datasets exist with domain gaps, causing synthetic data insufficiently aligned with the global distribution (\cref{fig:samples}). To tackle this gap, we fine-tune the vanilla SD with adaptive approaches~\cite{hu2021lora} using local data, which improves the alignment between the distributions of synthetic and real images. Finally, clients can train their models based on both real and synthetic images.

\subsection{Locally: Generating Synthetic Images}

\subsubsection{Alleviating Label Imbalance with Training-Free Approach}

Each client can directly use SD to generate synthetic images and alleviate label imbalance. Initially, we create text prompts like ``A photo of \{\texttt{class}\}'', where \{\texttt{class}\} represents each class in the global label set $\mathcal{Y}$. Using these prompts, clients can generate each training sample $s_i$ using the generative model $G$, defined as $s_i = G(z_i, p_i)$, where $z_i\sim\mathcal{N}(\mathbf{0}, \mathbf{I})$ is random noise, and $p_i$ represents each prompt. To elaborate, the sample is synthesized through an iterative denoising process starting from $z_i$, with $p$ providing conditional guidance. Then the predicted latent code is transformed into an image using the pre-trained VAE decoder~\cite{kingma2013auto}.

Finally, each client obtains their respective synthetic datasets, which are denoted as $\mathcal{S}_m=\{(s_i, y_i)\}^N_{i=1}$. We have the flexibility to adjust the quantity of synthetic images, allowing us to strike a balance between computational cost and performance.

\subsubsection{Fine-tuning for Better Adaptation to Local Data Domains}

Vanilla SD may lead to a domain gap. For instance, while real data consists of real-world images, synthetic data may mimic a painting style. Therefore, we are considering fine-tuning the SD based on the local data of clients using the adaptive approach LoRA~\cite{hu2021lora}. 

To ensure that the SD can generate synthetic samples that better align with the global data distribution, we intricately design the condition generation process for adaptation. Following~\cite{lei2023image}, we use the class name along with BLIP2~\cite{li2023blip} caption as the text prompt for each instance. While the text provides some adaptability, it overlooks essential visual information, including both low-level aspects like exposure and saturation, and high-level aspects such as object and scene co-occurrence. Visual information plays a crucial role in recovering global distribution, so we introduce the visual condition. Specifically, we utilize the CLIP~\cite{radford2021learning} Image Encoder to extract image features, calculate the mean embeddings for randomly selected images from each class, and subsequently estimate the intra-class mean feature distribution. Such mean feature is concatenated with the text embeddings to jointly create a condition, which is then injected into the cross-attention layer of UNet~\cite{ronneberger2015u} during the LoRA fine-tuning process. To summarize, our multi-modal condition takes the following form: ``A photo of \{\texttt{class}\}, \{\texttt{BLIP2 caption}\}, \{\texttt{intra-class visual feature}\}''.

\subsection{Globally: Aggregating Balanced Models}

Combining the real dataset $\mathcal{D}_m = \{(x_i, y_i)\}^{N_m}_{i=1}$ and synthetic dataset $\mathcal{S}_m=\{(s_i, y_i)\}^N_{i=1}$, each client has a balanced local dataset $\mathcal{O}_m=\mathcal{D}_m\bigcup\mathcal{S}_m$. We use FedAvg to conduct global generalization task. The server firstly dispatches a global model to the clients, and each client updates the model on $\mathcal{O}_m$, with the following objective:
\begin{equation}
\mathcal{L}=-\frac{1}{N}\sum\limits_{i=1}^N(y_i\log(\hat{y_i})+(1-y_i)\log(1-\hat{y_i})),
\label{eq:celoss}
\end{equation}
where $\hat{y_i}$ is the model prediction. After local training, each client obtains the local weights, \ie, $\vartheta_m$. Then, the server collects updated weights from clients, and aggregates the model $\vartheta$ with $\sum\limits_{m=1}^M\frac{N_m}{K}\vartheta_m$, where $M$ is the number of clients, and $K=\sum\limits_{m=1}^M N_m$. The above describes one round of communication, which can be repeated until it reaches the maximum rounds $T$. Finally, $\vartheta$ represents the ultimate model for the global generalization task.

During the local personalization task, each client obtains the global generalization parameter, \ie, $\vartheta$, and fine-tunes it on its local dataset $O_m$, with the same objective as in~\cref{eq:celoss}. Finally, they obtain the local parameters, \ie, $\vartheta_m$, respectively. The overall pseudocode of both generalization and personalization task can be seen in Appendix Algorithm~\ref{alg:code}.

\vspace{-4pt}

\section{Experiments}

\subsection{Experimental Setup}
\label{sec:setup}

\begin{figure*}[t!]
  \begin{center}
  \includegraphics[width=1.0\linewidth]{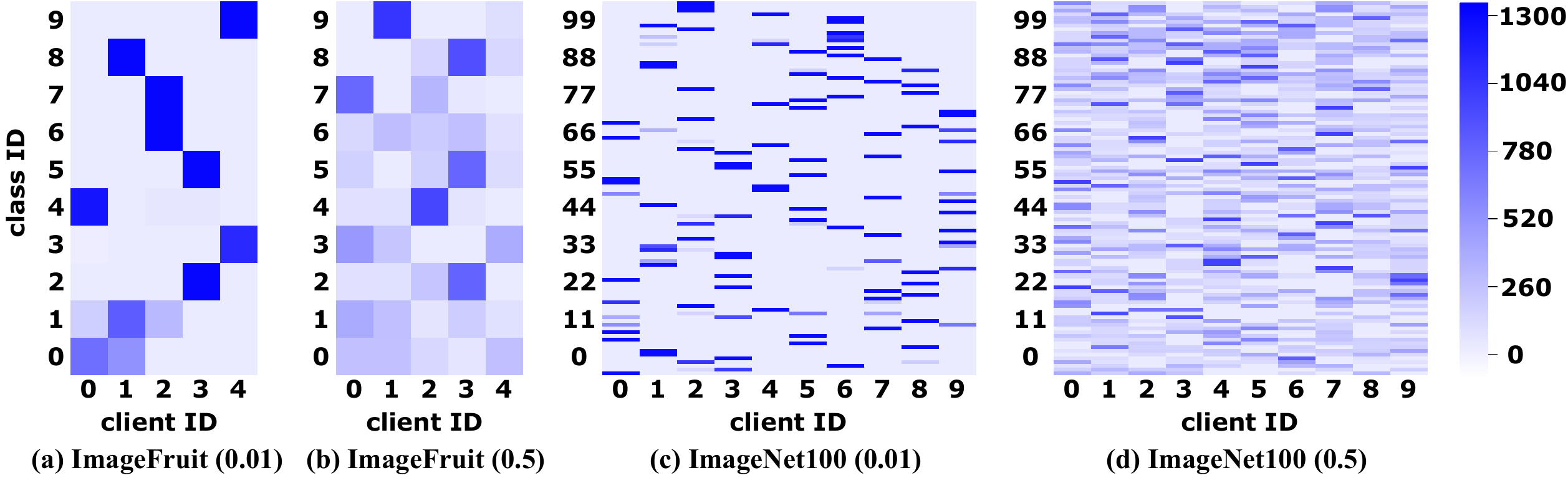}
  \caption{The data distributions of clients after data partition, where 0.01 and 0.5 are the $\beta$ values. The color bar shows the quantity of samples, and each rectangle represents the quantity of samples of a particular class in a client. Here we take 10-cateogry ImageFruit and 100-category ImageNet100 as examples.}
  \label{fig:partition}
  \end{center}
\end{figure*}

\begin{table*}[t!]
    \begin{center}
    \scalebox{0.67}{    
	\begin{tabular}{l|cc|cc|cc|cc|cc}
		\toprule
		\multirow{2}{*}{Method} &
        \multicolumn{2}{c|}{ImageFruit} &
        \multicolumn{2}{c|}{ImageNet100} & 
        \multicolumn{2}{c|}{CUB} &
        \multicolumn{2}{c|}{Cars} &
        \multicolumn{2}{c}{EuroSAT} \\
		& $\beta=0.01$ & $\beta=0.5$ & $\beta=0.01$ & $\beta=0.5$ & $\beta=0.01$ & $\beta=0.5$ & $\beta=0.01$ & $\beta=0.5$ & $\beta=0.01$ & $\beta=0.5$ \\
		\midrule
        Centralized$^{\text{\textdagger}}$ & \multicolumn{2}{c|}{$78.2_{\pm1.3}$} & \multicolumn{2}{c|}{$77.1_{\pm1.5}$} & \multicolumn{2}{c|}{$82.3_{\pm1.2}$} & \multicolumn{2}{c|}{$89.5_{\pm0.7}$} & \multicolumn{2}{c}{$95.6_{\pm1.1}$} \\
        \midrule
		FedAvg~\cite{mcmahan2017communication} & $29.0_{\pm2.0}$ & $51.2_{\pm1.9}$ & $36.3_{\pm1.5}$ & $44.6_{\pm1.8}$ & $42.2_{\pm2.7}$ & $59.6_{\pm1.6}$ & $49.5_{\pm1.7}$ & $63.8_{\pm1.5}$ & $56.2_{\pm1.3}$ & $72.9_{\pm1.8}$ \\
		FedProx~\cite{li2020federated} & $29.7_{\pm1.6}$ & $51.5_{\pm1.2}$ & $37.2_{\pm1.3}$ & $45.1_{\pm1.6}$ & $43.1_{\pm2.2}$ & $60.8_{\pm1.5}$ & $49.8_{\pm2.0}$ & $64.5_{\pm1.2}$ & $57.9_{\pm1.2}$ & $73.9_{\pm1.7}$ \\
		Scaffold~\cite{karimireddy2020scaffold} & $31.2_{\pm1.8}$ & $53.3_{\pm2.2}$ &  $40.6_{\pm1.7}$ & $49.3_{\pm1.3}$ & $44.9_{\pm2.0}$ & $62.3_{\pm1.8}$ & $52.7_{\pm1.3}$ & $66.1_{\pm1.1}$ & $58.8_{\pm1.5}$ & $74.2_{\pm1.7}$ \\
		FedNova~\cite{wang2020tackling} & $32.7_{\pm1.8}$ & $53.7_{\pm1.5}$ & $42.1_{\pm1.9}$ & $50.3_{\pm1.5}$ & $45.6_{\pm2.1}$ & $63.2_{\pm1.6}$ & $52.9_{\pm1.1}$ & $67.3_{\pm2.1}$ & $59.3_{\pm1.6}$ & $75.1_{\pm1.9}$ \\
        FedOpt~\cite{reddi2020adaptive} & $32.8_{\pm1.2}$ & $54.8_{\pm1.8}$ & $42.6_{\pm1.9}$ & $52.1_{\pm1.6}$ & $46.1_{\pm2.0}$ & $63.9_{\pm1.3}$ & $53.1_{\pm1.6}$ & $68.2_{\pm1.5}$ & $59.9_{\pm1.5}$ & $77.8_{\pm1.7}$ \\
		MOON~\cite{li2021model} & $33.6_{\pm1.9}$ & $55.3_{\pm1.1}$ & $43.2_{\pm1.5}$ & $52.9_{\pm1.2}$ & $47.6_{\pm2.1}$ & $65.2_{\pm2.6}$ & $54.6_{\pm1.4}$ & $69.8_{\pm1.3}$ & $61.0_{\pm1.8}$ & $79.2_{\pm2.1}$ \\ 
        FedAlign~\cite{mendieta2022local} & $37.2_{\pm1.3}$ & $58.0_{\pm1.3}$ & $46.9_{\pm1.6}$ & $56.1_{\pm1.3}$ & $51.8_{\pm2.3}$ & $67.6_{\pm0.8}$ & $55.2_{\pm0.9}$ & $72.2_{\pm1.3}$ & $62.6_{\pm0.9}$ & $81.8_{\pm1.6}$ \\
        DynaFed~\cite{pi2023dynafed} & $40.1_{\pm1.7}$ & $61.2_{\pm1.6}$ & $49.3_{\pm1.5}$ & $60.7_{\pm1.0}$ & $53.2_{\pm1.9}$ & $69.1_{\pm1.7}$ & $57.8_{\pm1.3}$ & $75.1_{\pm1.4}$ & $64.0_{\pm1.9}$ & $83.2_{\pm1.3}$ \\
		\midrule
		{\bf ReGL (TF)} & \underline{70.6}$_{\pm1.8}$ & \underline{72.6}$_{\pm1.3}$ & \underline{70.3}$_{\pm1.2}$ & \underline{71.3}$_{\pm1.6}$ & \underline{75.6}$_{\pm1.1}$ & \underline{78.2}$_{\pm1.9}$ & \underline{79.7}$_{\pm1.3}$ & \underline{81.2}$_{\pm1.5}$ & \underline{86.0}$_{\pm1.5}$ & \underline{87.6}$_{\pm1.1}$ \\
		{\bf ReGL (FT)} & \textbf{77.6}$_{\pm1.2}$ & \textbf{78.8}$_{\pm1.5}$ & \textbf{76.8}$_{\pm1.3}$ & \textbf{77.2}$_{\pm1.0}$ & \textbf{80.3}$_{\pm1.5}$ & \textbf{82.5}$_{\pm1.5}$ & \textbf{88.1}$_{\pm2.1}$ & \textbf{89.5}$_{\pm1.3}$ & \textbf{92.1}$_{\pm1.6}$ & \textbf{93.3}$_{\pm0.6}$ \\
		\bottomrule
	\end{tabular}
    }
    \end{center}
    \vspace{-5pt}
    \caption{{\bf Main Results I}: Global generalization performance on distribution-based label skew. The best results are highlighted in \textbf{bold}, and the second best is \underline{underlined}. ${\text{\textdagger}}$: We train a centralized model on the entire dataset without distributing data to multiple clients. \textbf{TF}: Using training-free generative model. \textbf{FT}: Fine-tuning the generative model adaptively.}
    \label{tab:main_result}
    \vspace{-0.3cm}
\end{table*}

\paragraph{Datasets and Data Partition} We conduct experiments on two challenging ImageNet~\cite{russakovsky2015imagenet} subsets: 10-category ImageFruit~\cite{zhang2023federated} and 100-category ImageNet100~\cite{tian2020contrastive}. We also test on three fine-grained datasets, named CUB~\cite{wah2011caltech}, Cars~\cite{krause20133d}, and satellite images EuroSAT~\cite{helber2019eurosat}.

In this paper, we use widely used Dirichlet distribution~\cite{yurochkin2019bayesian, li2021model} to simulate label distribution skew. Specifically, we sample $p_k$ from $Dir_N(\beta)$ and allocate a $p_{k,j}$ proportion of the instances of class $k$ to client $j$, where $N$ represents the client number, and $\beta$ controls the label imbalance, with a lower $\beta$ indicating a more skewed distribution. We evaluate under $\beta=0.5$ and $\beta=0.01$ (highly skewed). The distributions are detailed in~\cref{fig:partition}. We also consider an extremely challenging setting: a party with a single label~\cite{yu2020federated, li2022federated}, where clients have numerous missing classes. This also exists in real world, where we can employ FL to train a speaker recognition model, while each device only has its respective user data~\cite{li2022federated}.

Moreover, when evaluating personalization, the test set follows the partitioning of training set. Suppose there are $k_{train}$ samples per category in the training set, and $k_{test}$ in the test set, we randomly select test data from each category to maintain a consistent $k_{train}:k_{test}$ ratio between the training and test data for each category within each client.

\paragraph{Baselines and Implementation} We select typical non-IID methods as our baselines. For distribution-based skew, we allocate 5 clients for ImageFruit and EuroSAT; 10 for ImageNet100; and 20 for CUB and Cars. For skew with missing classes, we set client number equal to category number, with each client containing data from one class. 

We train a global ResNet34~\cite{he2016deep} model for generalization task. Then, each client fine-tunes their local model for personalization task, where we select the highest accuracy from each client and calculate the average. We conduct each experiment with 5 random seeds and report the average and standard deviation. More details can be found in~\cref{sec:more_details}.

\subsection{Comparisons with Baselines}
\label{sec:main_experiments}

\begin{wraptable}{R}{7.0cm}
\fontsize{7pt}{10pt}\selectfont
\begin{center}
\vskip -0.25in
\begin{tabular}{>{\arraybackslash}m{1.2cm} | >{\centering\arraybackslash}m{0.6cm}>{\centering\arraybackslash}m{0.7cm}>{\centering\arraybackslash}m{0.63cm}>{\centering\arraybackslash}m{0.63cm}>{\centering\arraybackslash}m{0.7cm}}
\toprule
Method & Fruit & IN100 & CUB & Cars & EuroSAT\\
        \midrule
        FedAvg~\cite{mcmahan2017communication} & $11.9_{\pm3.6}$ & $16.7_{\pm2.9}$ & $21.3_{\pm2.6}$ & $29.7_{\pm2.0}$ & $36.9_{\pm2.1}$ \\
		FedProx~\cite{li2020federated} & $12.2_{\pm3.3}$ & $18.2_{\pm3.1}$ & $23.9_{\pm2.1}$ & $31.9_{\pm2.5}$ & $37.3_{\pm1.8}$ \\
		Scaffold~\cite{karimireddy2020scaffold} & $12.9_{\pm2.8}$ & $20.0_{\pm2.0}$ & $24.2_{\pm2.1}$ & $33.2_{\pm2.2}$ & $38.2_{\pm1.3}$ \\
		FedNova~\cite{wang2020tackling} & $13.6_{\pm3.5}$ & $19.7_{\pm2.3}$ & $25.6_{\pm2.3}$ & $34.5_{\pm2.3}$ & $39.5_{\pm2.5}$ \\
        FedOpt~\cite{reddi2020adaptive} & $14.5_{\pm3.8}$ & $22.3_{\pm3.6}$ & $26.3_{\pm2.3}$ & $34.9_{\pm2.2}$ & $41.1_{\pm1.8}$ \\
		MOON~\cite{li2021model} & $14.8_{\pm2.9}$ & $24.1_{\pm3.3}$ & $28.9_{\pm1.9}$ & $35.8_{\pm1.7}$ & $42.9_{\pm2.0}$ \\
        FedAlign~\cite{mendieta2022local} & $19.8_{\pm2.2}$ & $27.8_{\pm3.0}$ & $31.2_{\pm1.7}$ & $37.2_{\pm1.9}$ & $43.9_{\pm2.2}$ \\
        DynaFed~\cite{pi2023dynafed} & $23.2_{\pm1.2}$ & $29.1_{\pm2.0}$ & $32.0_{\pm2.1}$ & $39.0_{\pm1.8}$ & $45.7_{\pm1.9}$ \\ 
		\midrule
		{\bf ReGL (TF)} & \underline{63.6}$_{\pm2.6}$ & \underline{68.8}$_{\pm2.2}$ & \underline{69.2}$_{\pm2.1}$ & \underline{75.8}$_{\pm2.3}$ & \underline{79.8}$_{\pm2.1}$ \\
		{\bf ReGL (FT)} & \textbf{72.5}$_{\pm3.0}$ & \textbf{73.2}$_{\pm1.9}$ & \textbf{75.1}$_{\pm1.7}$ & \textbf{83.2}$_{\pm1.3}$ & \textbf{88.1}$_{\pm1.6}$ \\
		\bottomrule
\end{tabular}
\end{center}
\vskip -0.08in
\caption{{\bf Main Results II}: Generalization performance when dealing with missing classes.}
\label{tab:missing_class_result}
\vskip -0.15in
\end{wraptable}

\paragraph{Performance on Generalization} As shown in~\cref{tab:main_result}, with training-free generation, our ReGL consistently outperforms all previous methods. When fine-tuning on local data, the performance is further enhanced, demonstrating the effectiveness of our paradigm. Notably, our adaptive ReGL surpasses FedAvg by significant margins of 48.6\% and 40.5\% on the ImageFruit and ImageNet100, with $\beta=0.01$. Furthermore, as data heterogeneity increases (\ie, with smaller $\beta$), our method exhibits greater performance gains compared to previous methods, highlighting the effectiveness in addressing label imbalance.

Moreover, previous works underperform the centralized model by about 35\% due to the data heterogeneity among clients, where local optima are generally far from the global optima. In contrast, we recover the global distribution, aligning the local and global optima. As shown in~\cref{tab:main_result}, our method achieves accuracies that match the centralized baselines.

\begin{wrapfigure}{r}{0.46\textwidth} 
\begin{center}
\vskip -0.3in
\centerline{\includegraphics[width=0.46\columnwidth]{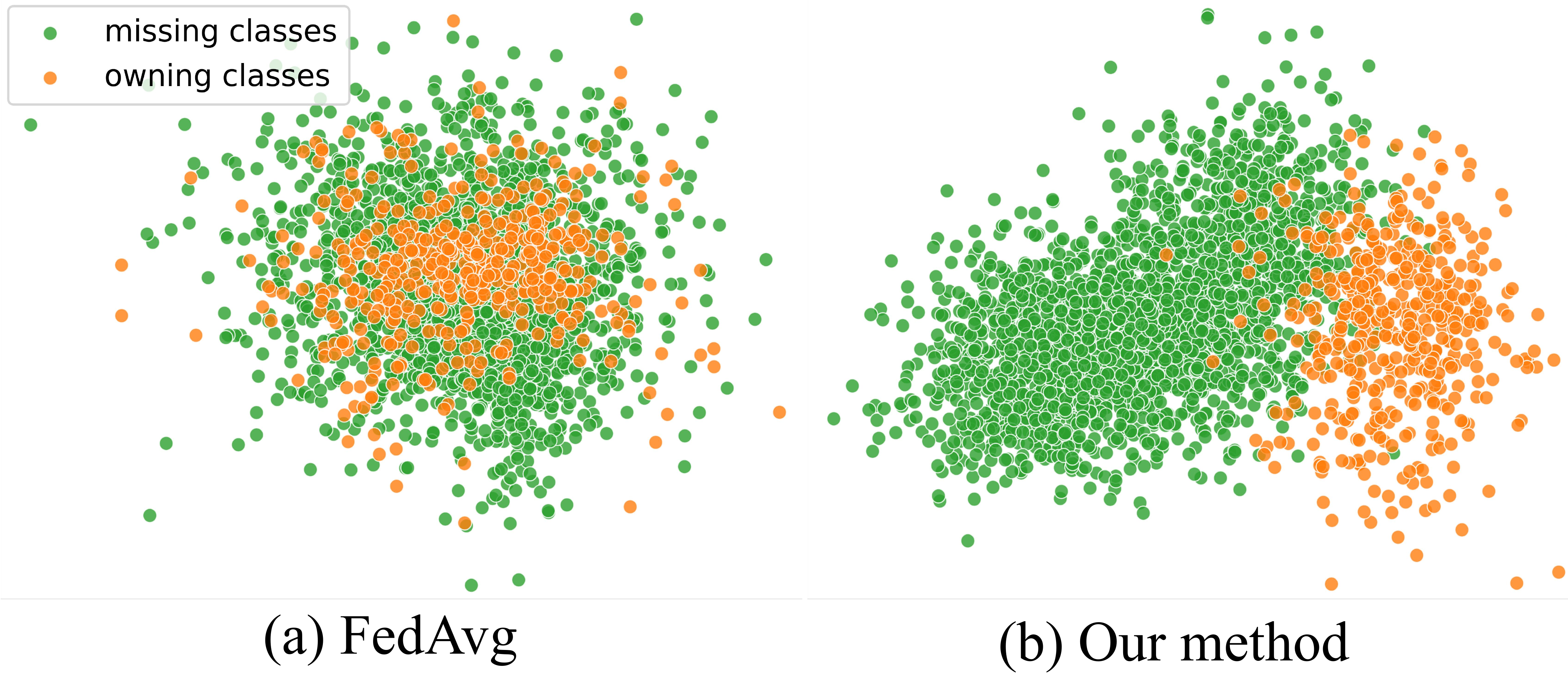}}
\vskip -0.1in
\caption{T-SNE visualization on owning and missing classes. (a) FedAvg mixes all samples indiscriminately, while (b) our method can effectively distinguish them.}
\label{fig:tsne_missing}
\end{center}
\vskip -0.4in
\end{wrapfigure}

\paragraph{Dealing with Missing Classes}~\cref{tab:missing_class_result} indicates that numerous missing classes lead to a catastrophic decline in accuracy for previous methods, which is primarily due to the significant disparity between local and global distributions. Most previous approaches rely on long-tail learning and struggle to effectively tackle the issue of missing classes. In contrast, our method demonstrates robustness by maintaining stable performance and achieving an accuracy approximately 40\% higher than that of previous methods. We further compare the T-SNE~\cite{van2008visualizing} visualization of our method and FedAvg on both owning and missing classes. As illustrated in~\cref{fig:tsne_missing}, after the update, FedAvg fails to distinguish between samples from owning and missing classes, resulting in poor performance. In contrast, our method leverages a generative model to complement the missing classes for each client, thus recovering each local distribution back to the global distribution and effectively learning discriminative features for samples of missing and owning classes.

\begin{table*}[t!]
    \begin{center}
    \scalebox{0.67}{    
	\begin{tabular}{l|cc|cc|cc|cc|cc}
		\toprule
		\multirow{2}{*}{Method} &
        \multicolumn{2}{c|}{ImageFruit} &
        \multicolumn{2}{c|}{ImageNet100} & 
        \multicolumn{2}{c|}{CUB} &
        \multicolumn{2}{c}{Cars} &
        \multicolumn{2}{c}{EuroSAT} \\
		& $\beta=0.01$ & $\beta=0.5$ & $\beta=0.01$ & $\beta=0.5$ & $\beta=0.01$ & $\beta=0.5$ & $\beta=0.01$ & $\beta=0.5$ & $\beta=0.01$ & $\beta=0.5$ \\
		\midrule
        Separate$^{\text{\textdagger}}$ & $56.8_{\pm2.1}$ & $22.3_{\pm1.2}$ & $59.6_{\pm1.6}$ & $23.6_{\pm0.9}$ & $62.3_{\pm1.6}$ & $30.6_{\pm1.1}$ & $66.7_{\pm1.5}$ & $35.2_{\pm1.8}$ & $83.2_{\pm1.2}$ & $47.5_{\pm1.2}$ \\
        \midrule
		FedAvg~\cite{mcmahan2017communication} & $33.5_{\pm2.1}$ & $42.6_{\pm2.3}$ & $38.6_{\pm2.8}$ & $43.1_{\pm1.9}$ & $43.8_{\pm2.2}$ & $50.7_{\pm1.5}$ & $47.3_{\pm2.1}$ & $54.6_{\pm2.3}$ & $53.9_{\pm2.6}$ & $65.2_{\pm1.7}$ \\
		FedRep~\cite{collins2021exploiting} & $58.1_{\pm1.9}$ & $41.9_{\pm2.4}$ & $60.0_{\pm2.1}$ & $43.7_{\pm2.6}$ & $66.1_{\pm2.1}$ & $50.8_{\pm2.0}$ & $70.2_{\pm2.3}$ & $57.5_{\pm1.9}$ & $77.2_{\pm2.2}$ & $65.9_{\pm1.9}$ \\
		FedBN~\cite{li2021fedbn} & $59.2_{\pm1.1}$ & $42.7_{\pm0.8}$ & $61.5_{\pm1.4}$ & $43.9_{\pm1.8}$ & $67.2_{\pm2.4}$ & $51.3_{\pm1.6}$ & $71.3_{\pm1.9}$ & $58.3_{\pm2.1}$ & $78.7_{\pm1.2}$ & $67.1_{\pm1.8}$ \\
		FedAMP~\cite{huang2021personalized} & $60.2_{\pm1.7}$ & $43.0_{\pm1.6}$ & $63.8_{\pm2.0}$ & $45.1_{\pm1.8}$ & $67.9_{\pm1.8}$ & $52.7_{\pm1.3}$ & $72.0_{\pm2.2}$ & $59.6_{\pm1.8}$ & $79.3_{\pm2.0}$ & $68.9_{\pm1.6}$ \\
        pFedGate~\cite{chen2023efficient} & $63.9_{\pm2.1}$ & $44.6_{\pm1.6}$ & $65.6_{\pm2.2}$ & $46.8_{\pm1.3}$ & $70.2_{\pm1.9}$ & $55.1_{\pm1.9}$ & $74.2_{\pm2.1}$ & $61.9_{\pm1.7}$ & $80.5_{\pm1.4}$ & $69.8_{\pm1.8}$ \\  
        FedCAC~\cite{wu2023bold} & $66.3_{\pm0.9}$ & $46.2_{\pm1.3}$ & $67.5_{\pm2.5}$ & $49.7_{\pm2.8}$ & $70.8_{\pm1.6}$ & $56.3_{\pm2.0}$ & $75.0_{\pm1.8}$ & $62.3_{\pm1.9}$ & $81.5_{\pm2.1}$ & $70.3_{\pm1.9}$ \\ 
		\midrule
		{\bf ReGL (TF)} & \underline{75.1}$_{\pm0.8}$ & \underline{73.9}$_{\pm0.9}$ & \underline{74.9}$_{\pm1.2}$ & \underline{73.5}$_{\pm1.2}$ & \underline{80.8}$_{\pm1.8}$ & \underline{79.2}$_{\pm1.1}$ & \underline{85.6}$_{\pm1.9}$ & \underline{83.1}$_{\pm1.6}$ & \underline{89.2}$_{\pm1.6}$ & \underline{88.3}$_{\pm1.5}$ \\
        {\bf ReGL (FT)} & \textbf{81.6}$_{\pm1.2}$ & \textbf{79.7}$_{\pm1.5}$ & \textbf{80.1}$_{\pm1.3}$ & \textbf{78.9}$_{\pm1.0}$ & \textbf{83.6}$_{\pm1.9}$ & \textbf{82.9}$_{\pm2.0}$ & \textbf{92.2}$_{\pm1.8}$ & \textbf{91.0}$_{\pm1.3}$ & \textbf{95.2}$_{\pm1.1}$ & \textbf{94.5}$_{\pm0.9}$ \\
		\bottomrule
	\end{tabular}
    }
    \end{center}
    \vspace{-4pt}
    \caption{{\bf Main Results III}: Local personalization performance on distribution-based label skew. ${\text{\textdagger}}$: Each model is solely trained on local data without cross-client collaboration.}
    \label{tab:personalized}
    \vspace{-6pt}
\end{table*}

\paragraph{Performance on Personlization} We also evaluate the personalization of our ReGL by comparing it with several state-of-the-art Personalized FL methods. As shown in~\cref{tab:personalized}, we can observe that most of the state-of-the-art methods outperform the Separate baseline, especially with $\beta=0.5$, which highlights the importance of collaboration among various clients. Additionally, our method outperforms all previous approaches across all datasets. For instance, with $\beta=0.01$, our method surpasses previous methods by an average of 15\% across five datasets and by over 30\% with $\beta=0.5$. We attribute this result to two key aspects: 1) Excellent personalization is inherently based on robust generalization. As demonstrated in~\cref{tab:main_result}, our method achieves the best global generalization, ensuring that each client has a solid foundation for personalized fine-tuning. 2) During fine-tuning, each client can utilize both real and synthetic data. The synthetic data conforms to the same distribution as the real data, significantly augmenting the training dataset.

\subsection{Ablation Studies and More Analyses of Our ReGL}
\label{sec:ablation_and_analyses}

We conduct comprehensive ablation studies and analysis on the validation sets of ImageFruit and ImageNet100, and focus on the global generalization evaluation.   

\begin{wrapfigure}{r}{0.5\textwidth} 
\begin{center}
\vskip -0.3in
\centerline{\includegraphics[width=0.5\columnwidth]{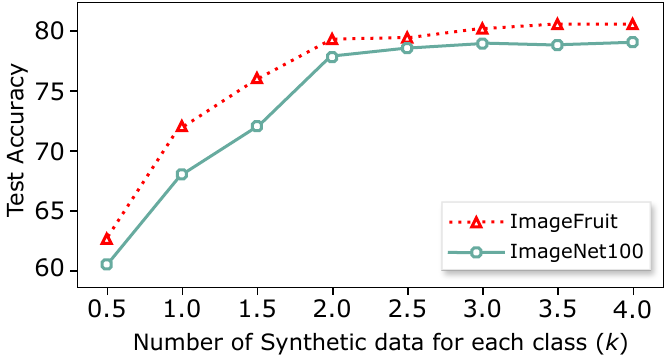}}
\vskip -0.1in
\caption{Performance with various synthetic data volume.}
\label{fig:syn_num}
\end{center}
\vskip -0.4in
\end{wrapfigure}

\paragraph{Number of Synthetic Data} As shown in~\cref{fig:syn_num}, we investigate the effect of synthetic image volume on performance with $\beta=0.5$. We generate different numbers of images per class for each client, ranging from 0.5$k$ to 4$k$. Obviously, as the number of synthetic data increases, the accuracy consistently improves. For example, increasing the number of synthetic data from 0.5$k$ to 2$k$ results in an accuracy improvement, from 60\% to 77\%, on the ImageNet100. The accuracy can further improve with more synthetic images, such as 4$k$.  In this way, our method can significantly enhance performance
by generating more training samples without the need for complex design. Considering both efficiency and performance, we set the synthetic data volume per class for each client to 2$k$.

\begin{wrapfigure}{r}{0.5\textwidth} 
\begin{center}
\vskip -0.1in
\centerline{\includegraphics[width=0.5\columnwidth]{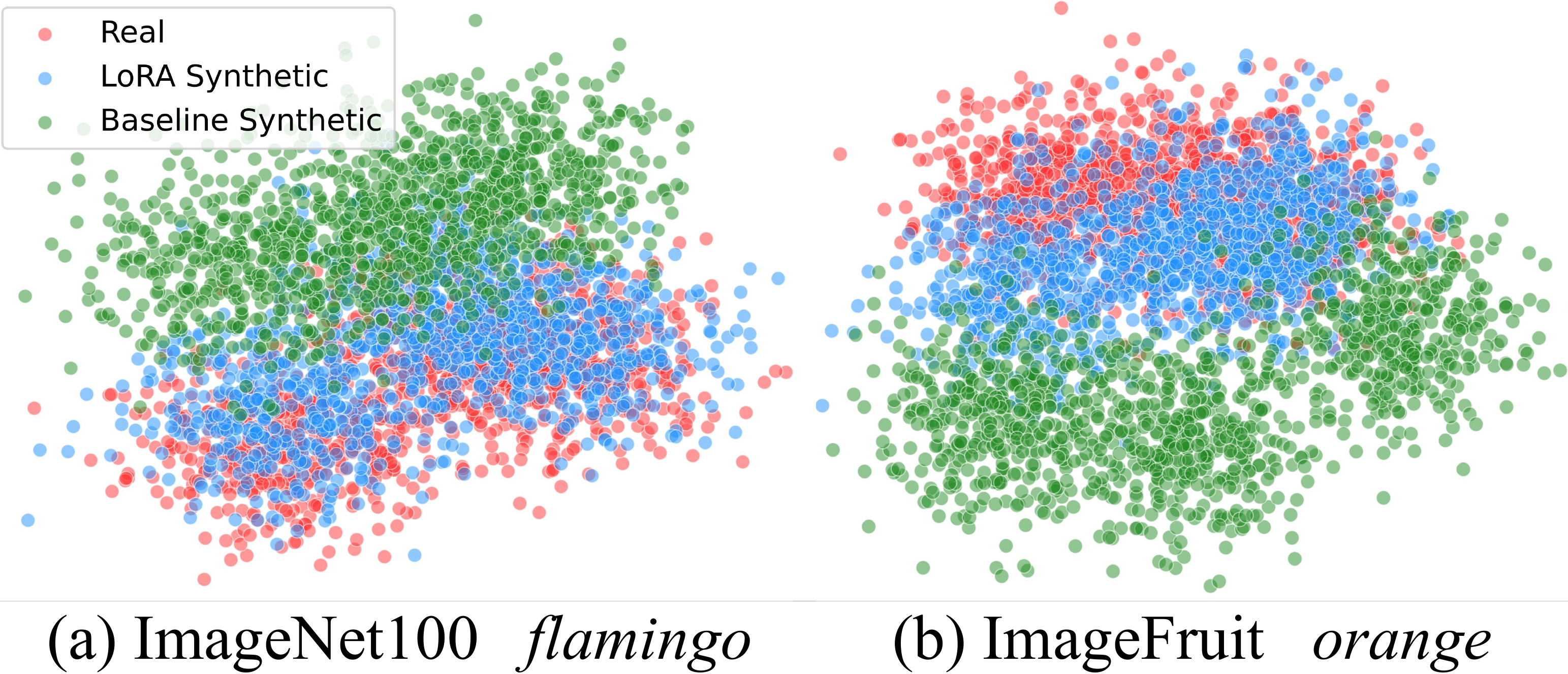}}
\vskip -0.1in
\caption{T-SNE visualization of the data distribution. Our LoRA synthetic images are much closer to the global distribution.}
\label{fig:tsne}
\end{center}
\vskip -0.4in
\end{wrapfigure}

\paragraph{Recovering Global Data Distribution} Our method essentially recovers the heterogeneous local distributions to match the global distribution, thereby eliminating label distribution skew. Here, we verify the data distribution from two perspectives: {\bf 1)} We visually compare real images, training-free baseline synthetic images, and adaptive LoRA synthetic images from ImageFruit. As shown in~\cref{fig:samples}, although the synthetic images generated by vanilla SD have correct class information, their styles are very different from the real images. In other words, the global distribution is only partially recovered. In contrast, our LoRA synthetic images have a style similar to the real images, leading to a well-recovered data distribution. {\bf 2)} We compare the t-SNE visualization of these distributions, which is presented in~\cref{fig:tsne}. We can observe a significant misalignment between the baseline synthetic and real data distributions. In contrast, the distribution of LoRA synthetic images closely match the real distribution, validating the recovery of the global distribution by our method.

\begin{wraptable}{R}{5.9cm}
\fontsize{7pt}{8pt}\selectfont
\begin{center}
\vskip -0.25in
\begin{tabular}{>{\arraybackslash}m{0.8cm} | >{\centering\arraybackslash}m{0.35cm}>{\centering\arraybackslash}m{0.35cm}>{\centering\arraybackslash}m{0.35cm}>{\centering\arraybackslash}m{0.35cm}>{\centering\arraybackslash}m{0.35cm}| >{\centering\arraybackslash}m{0.35cm}}
    \toprule
    \multirow{2}{*}{Methods} & \multicolumn{5}{c|}{Degree of Skewness $\beta$} & \multirow{2}{*}{\,IID\,}\\
    & $\,0.01\,$ & $\,0.05\,$ & $\,0.1\,$ & $\,0.3\,$ & $\,0.5\,$ & \\
    \midrule
    FedAvg & 30.1 & 33.6 & 39.7 & 49.2 & 52.1 & 73.6 \\
    FedProx & 30.8 & 33.8 & 40.7 & 50.3 & 52.6 & 74.0 \\
    FedNova & 31.9 & 34.7 & 41.3 & 50.5 & 54.0 & 74.2 \\
    FedOpt & 32.6 & 35.2 & 42.9 & 51.6 & 55.1 & 75.1 \\
    MOON & 33.7 & 36.9 & 43.2 & 51.9 & 55.9 & 76.6 \\
    Ours & {\bf 77.3} & {\bf 77.8} & {\bf 78.3} & {\bf 78.9} & {\bf 79.2} & {\bf 81.0} \\
    \bottomrule
\end{tabular}
\end{center}
\vskip -0.08in
\caption{Performance under various skewness. $\beta=0.01$ indicates highly skewed data, while IID represents no skewness.}
\label{tab:ablation_more_beta}
\vskip -0.1in
\end{wraptable}

\paragraph{Robustness on Highly Skewed Data} To assess the robustness of ReGL on highly skewed data, we conduct experiments on ImageFruit, wherein $\beta$ varies from $0.01$ to $0.5$. Additionally, we test the models in an IID (Independent and Identically Distributed) setting, where the data is evenly distributed among the clients. As shown in~\cref{tab:ablation_more_beta}, our method outperforms previous methods by around 15\% under the IID setting, which can be attributed to the generation of synthetic images, serving as data augmentation. As the skewness increases, our method remains robust, with performance consistently exceeding 77\%, whereas other methods experience a sharp decline in performance, dropping from 75\% (IID) to 30\% ($\beta=0.01$). This demonstrates that our method can significantly improve performance in scenarios of extreme label distribution skew.

\paragraph{More Ablation Studies and Analyses} Please find more details in~\cref{sec:more_ablation_studies}, including 1) Local Epochs $E_{local}$, 2) Number of Clients $M$, 3) Analysis of Class-level Accuracy.

\section{Conclusion}

In this paper, we propose a novel approach named \textbf{ReGL} for label distribution skew in FL, which aims to \textbf{Re}cover the \textbf{G}lobal data distribution \textbf{L}ocally. Our approach leverages foundation generative models to generate synthetic data for each client, filling in the gaps for minority and missing classes while approximating the global distribution. Additionally, we adaptively fine-tune the generative models using local data to better align the local distribution with the global one. Comprehensive experiments show that our ReGL effectively tackles label distribution skew, outperforming previous state-of-the-art methods.
\clearpage
\appendix

\section{Detailed Related Work}
\label{sec:detail_related_work}

Label distribution skew poses a challenge in training a model to perform well across all clients. Numerous works have attempted to address label imbalance, which can be mainly categorized into four groups: {\bf 1) Incorporate Momentum and Adaptive Methods.} Many recent works try to incorporate optimization methods into FL. For example, FedOpt~\cite{reddi2020adaptive} consists of ServerOpt for the server and ClientOpt for the clients, which are used to update the global and local models, respectively. Both ClientOpt and ServerOpt can be set as any momentum and adaptive optimizers to enhance performance. {\bf 2) Reduce the Bias in Local Model Updates.} When dealing with heterogeneous data, local updates can introduce bias into the convergence process. Therefore, some methods focus on reducing this bias. SCAFFOLD~\cite{karimireddy2020scaffold} effectively employs control variates, a technique aimed at reducing variance, to correct client-drift within its local updates. {\bf 3) Regularize Local Objective Functions.} Some works attempt to penalize local models that deviate significantly from the global model by applying regularization to the local objectives. For instance, FedProx~\cite{li2020federated} employs the Euclidean distance between local and global models as a regularization function to prevent local models from drifting towards their respective local minima. {\bf 4) Consider Alternative Aggregation Methods.} The weighting aggregation determines the ultimate convergence point of the global model. Therefore, recent methods try to design an effective weighting scheme. For example, FedNova~\cite{wang2020tackling} optimizes the number of epochs in local updates and introduces a normalized averaging scheme to eliminate inconsistencies in objectives.

However, such optimization methods cannot address the fundamental issue of data distribution heterogeneity, meaning they cannot achieve truly outstanding performance. Moreover, another solution involves generating synthetic data from distributed data sources, as discussed in previous studies~\cite{hardy2019md, rasouli2020fedgan, mugunthan2021bias}. However, these methods cannot generate high-quality data or prevent potential privacy leakage, resulting in subpar performance. Different from the previous approaches, we start by recovering the global data distribution at the local level, thus aiming to fundamentally address the label imbalance in FL.

\section{More Implementation Details}
\label{sec:more_details}

We implement all methods using PyTorch. Before training, we utilize Stable Diffusion to generate synthetic images for each client based on the prompt ``A photo of \{\texttt{class}\}, real world images, high resolution''. In terms of data volume, for each class, we generate 1.3$k$ synthetic images for the ImageFruit and ImageNet100 datasets, 100 synthetic images for the CUB and Cars datasets, and 3$k$ synthetic images for the EuroSAT dataset. For the adaptive fine-tuning approach, we employ LoRA to fine-tune the Stable Diffusion within each client using their local data and set the $\alpha$ of U-Net to 0.8.

During the global generalization task, the batch size is set to 128, and the round of communication is set to 200. In each communication round, every client updates their weights for 5 epochs using the SGD optimizer. During the local personalization task, we select synthetic data for each client based on the categories of their real data. We fine-tune the global model at each client using their local data for 50 epochs with the SGD optimizer, resulting in a personalized local model for each client.

\begin{algorithm}
\caption{\small Pytorch-like Pseudocode of Our ReGL.}
\label{alg:code}
\definecolor{codeblue}{rgb}{0.25,0.5,0.5}
\definecolor{codekw}{rgb}{0.85, 0.18, 0.50}
\begin{lstlisting}[language=python]
datasets = []
SD = vanilla_Stable_Diffusion()
for client in all_clients:
    # use the real images of each client to fine tune the SD individually
    adaptive_SD = fine_tuning(real_images, SD) 
    # use adaptive SD to generate synthetic images
    syn_images = adaptive_SD.generation(prompts, noise)
    # package the real and synthetic images
    all_data = aggregation(real_images, syn_images)
    datasets.append(all_data)

# global generalization task
global_model = network()
for com in Rounds:
    local_weights = []
    # randomly select clients
    for client in selected_clients:
        # SGD update on real and synthetic images
        data = datasets[client]
        weights = local_update(data, global_model)
        local_weights.append(weights)
    # updating the global model by average
    global_model = model_aggregation(local_weights)

# local personalization task
local_models = []
for client in all_clients:
    # each client fine-tune the model for their personalized tasks
    data = datasets[client]
    local_weight = local_update(data, global_model)
    local_models.append(local_weight)

return global_model, local_models
\end{lstlisting}
\end{algorithm}

\section{More Ablation Studies and Analyses}
\label{sec:more_ablation_studies}

\begin{table}[t!]
    \begin{center}
    \scalebox{0.9}{    
	\begin{tabular}{c|l|cc|cc}
		\toprule
		\multirow{2}{*}{$E_{local}$} &
        \multirow{2}{*}{Methods} &
		\multicolumn{2}{c|}{ImageFruit} &
        \multicolumn{2}{c}{ImageNet100} \\
		& & $\beta=0.01$ & $\beta=0.5$ & $\beta=0.01$ & $\beta=0.5$ \\
		\midrule
        \multirow{4}{*}{1} & FedAvg & 23.7 & 39.2 & 28.8 & 33.6 \\
        & FedNova & 27.9 & 40.6 & 32.9 & 37.7 \\
        & MOON & 28.5 & 41.8 & 31.8 & 38.5\\
        & Ours & 60.7 & 65.2 & 60.1 & 63.2\\
        \midrule
        \multirow{4}{*}{5} & FedAvg & 30.1 & 52.1 & 37.0 & 43.9 \\
        & FedNova & 31.9 & 54.0 & 43.7 & 51.3 \\
        & MOON & 33.7 & 55.9 & 45.6 & 53.8 \\
        & Ours & \textbf{77.3} & \underline{79.2} & \underline{77.2} & \textbf{78.6} \\
        \midrule
        \multirow{4}{*}{10} & FedAvg & 31.9 & 52.7 & 36.5 & 42.8 \\
        & FedNova & 32.3 & 53.5 & 43.9 & 52.0 \\
        & MOON & 33.1 & 54.6 & 47.0 & 53.1 \\
        & Ours & 76.8 & \textbf{80.2} & \textbf{78.2} & \underline{77.9}\\
        \midrule
        \multirow{4}{*}{20} & FedAvg & 30.6 & 50.6 & 34.1 & 43.1 \\
        & FedNova & 31.8 & 52.3 & 43.6 & 53.5\\
        & MOON & 30.9 & 51.6 & 46.2 & 54.1 \\
        & Ours & \underline{77.0} & 78.6 & 76.0 & 77.2\\
		\bottomrule
	\end{tabular}
	}
    \end{center}
    \caption{Performance comparison of different local epochs.}
    \label{tab:ablation_local_epoch}
\end{table} 

\begin{table}[t!]
    \begin{center}
    \scalebox{0.9}{    
	\begin{tabular}{l|ccc|ccc}
		\toprule
        \multirow{2}{*}{Methods} &
		\multicolumn{3}{c|}{ImageFruit} &
        \multicolumn{3}{c}{ImageNet100} \\
		& $M=5$ & $M=50$ & $M=100$ & $M=10$ & $M=50$ & $M=100$ \\
		\midrule
        FedAvg & 30.1 & 14.5 & 9.1 & 37.0 & 20.5 & 13.3 \\
        FedProx & 30.7 & 14.9 & 9.6 & 38.8 & 20.8 & 14.1 \\
        FedNova & 31.9 & 18.8 & 11.7 & 43.7 & 22.6 & 14.9 \\
        FedOpt & 32.8 & 19.6 & 11.9 & 44.1 & 23.0 & 16.8 \\
        MOON & 33.7 & 20.2 & 12.7 & 45.6 & 25.9 & 17.1 \\
        Ours & {\bf 77.3} & {\bf 77.0} & {\bf 76.2} & {\bf 77.2} & {\bf 76.8} & {\bf 75.0} \\
		\bottomrule
	\end{tabular}
    }
    \end{center}
    \caption{Performance comparison of different number of clients.}
    \label{tab:ablation_number_of_client}
\end{table} 

\begin{figure}[t!]
  \centering
  \includegraphics[width=1.0\linewidth]{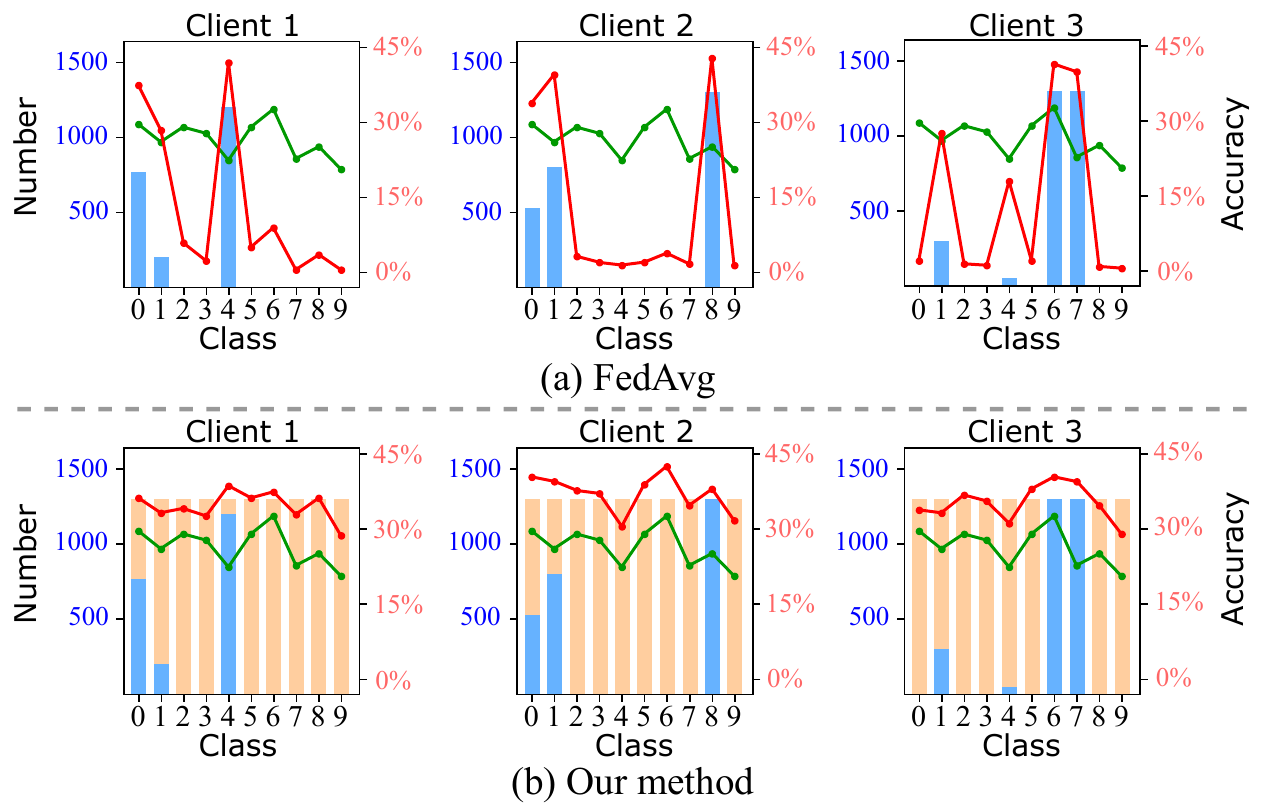}
  \caption{Class-level accuracy of FedAvg and our method on the skewed ImageFruit dataset. The \textcolor[RGB]{51, 153, 255}{blue histogram} and the \textcolor[RGB]{255, 153, 51}{orange histogram} represent the number of real and synthetic images for each class, respectively. The \textcolor[RGB]{0, 153, 0}{green line} and the \textcolor[RGB]{255, 0, 0}{red line} indicate the accuracy of each class before and after a certain local update, respectively.}
  \label{fig:class}
\end{figure}

\begin{figure*}[t!]
  \begin{center}
  \includegraphics[width=1.0\linewidth]{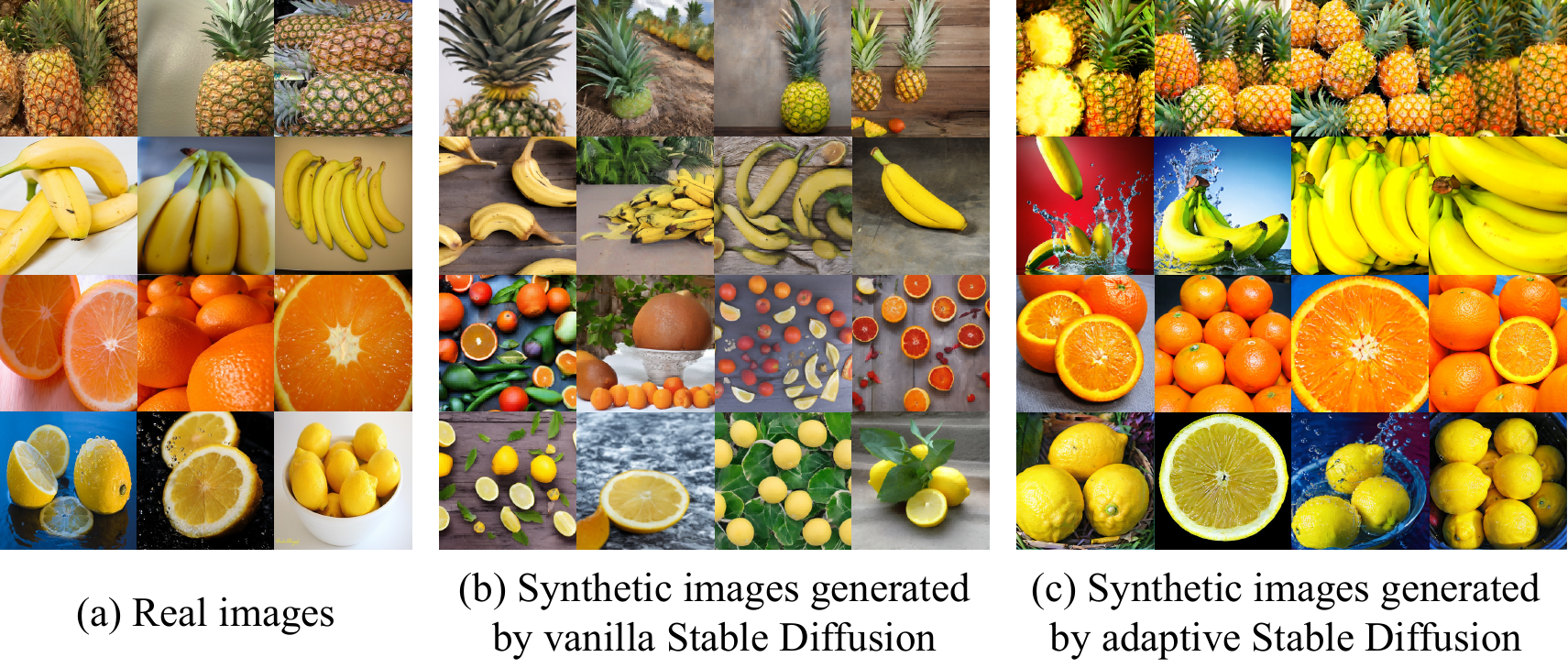}
  \caption{Visualization of real and synthetic images in ImageFruit dataset. While there is a significant difference between the real images and synthetic images generated by vanilla Stable Diffusion, adaptive approach can generate synthetic images in the style of real images.}
  \label{fig:samples}
  \vspace{-6pt}
  \end{center}
\end{figure*}

\paragraph{Local Epochs} Here, we increase the computation load per client in each round by expanding the number of local epochs, which we denote as $E_{local}$. We conduct numerous experiments on ImageFruit and ImageNet100 datasets, and compare our ReGL with previous algorithms in~\cref{tab:ablation_local_epoch}. It is evident that, when $E_{local}$ is set to 1, the performance of all methods degrades significantly. In this scenario, the number of local updates is too small, resulting in inadequate model training. Nonetheless, our method continues to exhibit competitive performance, achieving an accuracy of 60.7\% on the ImageFruit dataset and 60.1\% on the ImageNet100 dataset with $\beta=0.01$. As we increase the value of $E_{local}$, the performance of all methods generally improves. However, excessively large values of $E_{local}$ can lead to overfitting. Therefore, the optimal choice for our method is 5 epochs per round.

\paragraph{Number of Clients} To analyze the effect of the number of clients on performance, we train these methods with different numbers of clients $M$ on ImageFruit and ImageNet100 datasets. Specifically, we set $M=\{5, 50, 100\}$ for the ImageFruit and $M=\{10, 50, 100\}$ for the ImageNet100, with a skew degree of $\beta = 0.01$. As demonstrated in~\cref{tab:ablation_number_of_client}, when $M$ increases, the performance of all previous methods experiences a significant decline. This decline is especially pronounced when $M=100$, where the accuracy of previous methods drops to approximately 10\% on ImageFruit and 15\% on ImageNet100. We conjecture that as the number of clients increases, there are more skewed local models, leading to a poorer aggregated model. However, our method remains robust as $M$ increases, with almost no performance degradation. Based on these results, we can conclude that our method is highly suitable for scenarios involving a large number of clients. In cases where the number of clients is particularly high, for instance, exceeding 100, our method outperforms the previous approaches by about 60\%.

\paragraph{Analysis of Class-level Accuracy} Label distribution skew can lead to class inconsistency across clients, so we compare the accuracy of FedAvg and our method for each class before and after a certain local update on the ImageFruit dataset using the $\beta=0.01$ setting. \emph{For the sake of display, we generate synthetic images for each class in our method to ensure that the total number of real and synthetic images is 1.3$k$, which is slightly different from the settings in our main experiment.} The performance is shown in~\cref{fig:class}. Here, all local models on the test set have the same test accuracy before local updates, because these local models are equivalent to the global model. First, we can observe that in FedAvg, the local model is restricted to learning samples solely from the majority classes, leading to a sharp decline in accuracy for the remaining classes. This indicates that label distribution skew can lead to a biased model, severely impacting global model performance. While in our method, we eliminate label skew by generating synthetic images for each class, which enables a more effective local model update. The test accuracy of each class improves after local updates, \emph{resulting in a more robust federated learning system under label distribution skew.}

\bibliography{egbib}
\end{document}